\documentclass[10pt,twocolumn,letterpaper]{article}

\usepackage{cvpr}              % To produce the CAMERA-READY version
\usepackage{graphicx}
\usepackage{amsmath}
\usepackage{amssymb}
\usepackage{booktabs}
\usepackage{makecell}
\usepackage{stfloats}
\usepackage{float}
\usepackage{bbding}
\usepackage[accsupp]{axessibility}

\usepackage{multirow}%提供跨列命令\multicolumn{}{}{}

\usepackage[pagebackref,breaklinks,colorlinks]{hyperref}

\usepackage[capitalize]{cleveref}
\crefname{section}{Sec.}{Secs.}
\Crefname{section}{Section}{Sections}
\Crefname{table}{Table}{Tables}
\crefname{table}{Tab.}{Tabs.}

\def\cvprPaperID{2575} % *** Enter the CVPR Paper ID here
\def\confName{CVPR}
\def\confYear{2023}

\begin{document}

\title{A2J-Transformer: Anchor-to-Joint Transformer Network for 3D Interacting Hand Pose Estimation from a Single RGB Image}

\newcommand{\wcl}[1]{\textcolor{black}{#1}}

\author{Changlong~Jiang$^1$, Yang~Xiao$^1$$^\dag$, Cunlin~Wu$^1$, Mingyang~Zhang$^2$, Jinghong~Zheng$^1$, Zhiguo~Cao$^1$, \\ and Joey Tianyi~Zhou$^{3,4}$\\
\normalsize{$^1$Key Laboratory of Image Processing and Intelligent Control, Ministry of Education, School of Artificial} \\\normalsize{Intelligence and Automation, Huazhong University of Science and Technology, Wuhan 430074, China}\\
\normalsize{$^2$ Alibaba Group}\\
\normalsize{$^3$ Centre for Frontier AI Research (CFAR), Agency for Science, Technology and Research (A*STAR), Singapore}\\
\normalsize{$^4$Institute of High Performance Computing (IHPC), Agency for Science, Technology and Research (A*STAR), Singapore}\\
\tt\small {changlongj, Yang$\_$Xiao, cunlin$\_$wu}@hust.edu.cn,
changhai.zmy@alibaba-inc.com,\\
\tt\small {deepzheng,zgcao}@hust.edu.cn, 
\tt\small zhouty@cfar.a-star.edu.sg}

\maketitle
\let\thefootnote\relax\footnotetext{\dag Yang Xiao is corresponding author(Yang$\_$Xiao@hust.edu.cn).}

%%%%%%%%% ABSTRACT
\begin{abstract}
   3D interacting hand pose estimation from a single RGB image is a challenging task, due to serious self-occlusion and inter-occlusion towards hands, confusing similar appearance patterns between 2 hands, ill-posed joint position mapping from 2D to 3D, etc.. To address these, we propose to extend A2J-the state-of-the-art depth-based 3D single hand pose estimation method-to RGB domain under interacting hand condition. 
   Our key idea is to equip A2J with strong local-global aware ability to well capture interacting hands' local fine details and global articulated clues among joints jointly.
   To this end, A2J is evolved under Transformer's non-local encoding-decoding framework to build A2J-Transformer.
   It holds 3 main advantages over A2J. First, self-attention across local anchor points is built to make them global spatial context aware to better capture joints' articulation clues for resisting occlusion. 
   Secondly, each anchor point is regarded as learnable query with adaptive feature learning for facilitating pattern fitting capacity, instead of having the same local representation with the others. 
   Last but not least, anchor point locates in 3D space instead of 2D as in A2J, to leverage 3D pose prediction. Experiments on challenging InterHand 2.6M demonstrate that, A2J-Transformer can achieve state-of-the-art model-free performance (3.38mm MPJPE advancement in 2-hand case) and can also be applied to depth domain with strong generalization. The code is avaliable at \url{https://github.com/ChanglongJiangGit/A2J-Transformer}.

\end{abstract}

%--------------------------------------------------------------------------------------
%%%%%%%%% BODY TEXT
\section{Introduction}
\label{sec:intro}

\begin{figure}
%\vspace{-0.2cm}
%\setlength{\abovecaptionskip}{5 cm}
\centering
\includegraphics[height=3.8cm]{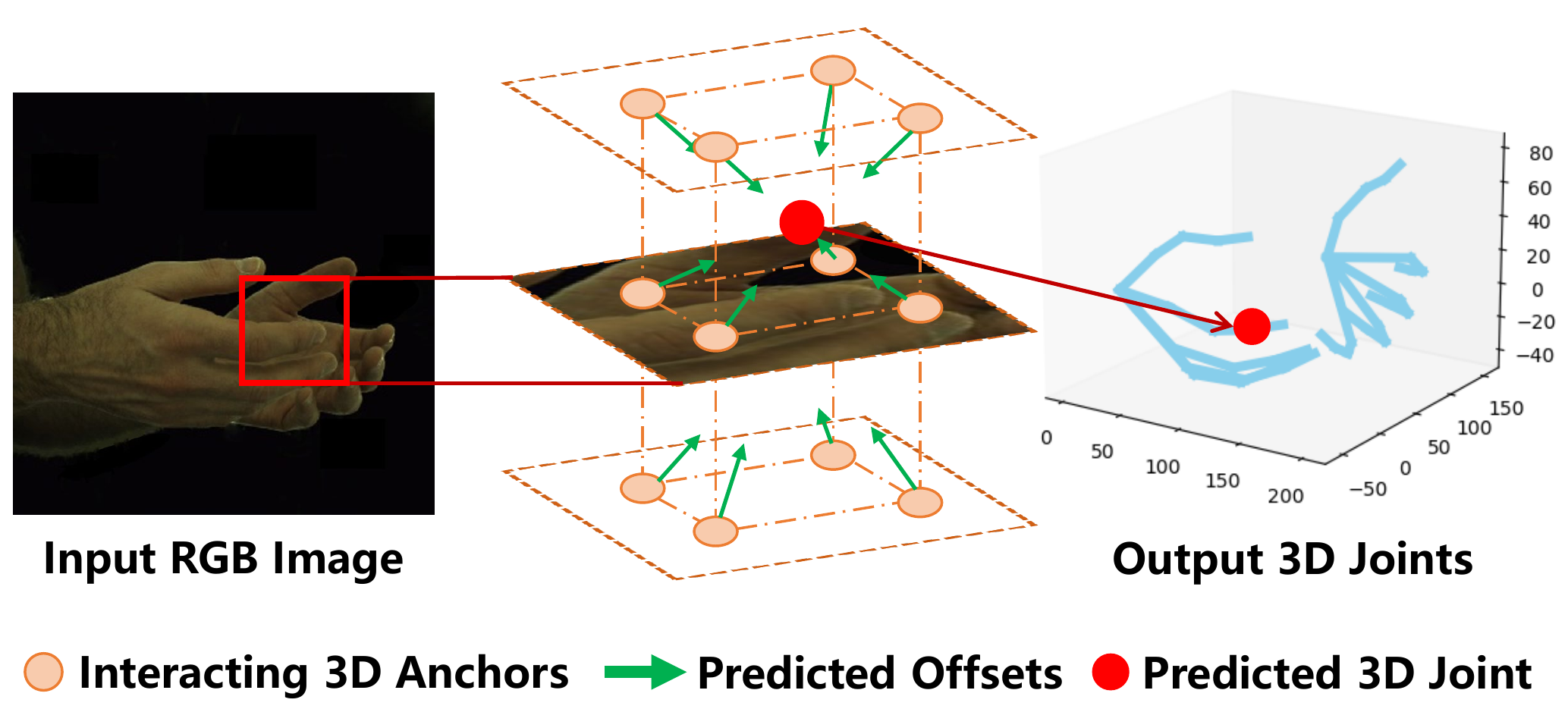}
\vspace{-8pt}
\caption{The main idea of A2J-Transformer. 3D anchors are uniformly set and act as local regressors to predict each hand joint. Meanwhile, they are also used as queries, and the interaction among them is established to acquire global context.}
% \vspace{-12pt}
\label{fig:figure1}
\vspace{-8pt}
\end{figure}

3D interacting hand pose estimation from a single RGB image can be widely applied to the fields of virtual reality, augmented reality, human-computer interaction, etc..~\cite{romero2009monocular,shotton2011real,tang2014latent}. Although the paid efforts, it still remains as a challenging research task due to the main issues of serious self-occlusion and inter-occlusion towards hands~\cite{huang2021survey,lin2021two,fan2021learning,hampali2022keypoint,meng20223d}, confusing similar appearance patterns between 2 hands~\cite{kim2021end,hampali2022keypoint,meng20223d}, and the ill-posed characteristics of estimating 3D hand pose via monocular RGB image~\cite{moon2018v2v,huang2021survey,fan2021learning}.

The existing methods can be generally categorized into model-based~\cite{tzionas2016capturing,smith2020constraining,li2022interacting,zhang2021interacting,moon2020interhand2,ballan2012motion,wang2020rgb2hands,oikonomidis2012tracking,cai2018weakly} and model-free~\cite{meng20223d,iqbal2018dual,xiong2019a2j,kim2021end,liu2021hand,cheng2021handfoldingnet,hampali2022keypoint,fan2021learning,lin2021two,moon2020interhand2} groups. 
Due to model's strong prior knowledge on hands, the former paradigm is overall of more promising performance. However, model-based methods generally require complex personalized model calibration, which is sensitive to initialization and susceptible to trap in local minima~\cite{hampali2022keypoint,hampali2020honnotate}. This is actually not preferred by the practical applications. Accordingly, we focus on model-free manner in regression way. The key idea is that, \emph{for effective 3D interacting hand pose estimation the predictor should be well aware of joints' local fine details and global articulated context simultaneously} to resist occlusion and confusing appearance pattern issues. To this end, we propose to extend the SOTA depth-based single hand 3D pose estimation method A2J~\cite{xiong2019a2j} to 3D interacting hand pose estimation task from a single RGB image.

Although A2J's superiority with ensemble local regression, intuitively applying it to our task cannot ensure promising performance, since it generally suffers from 3 main defects as below. First, the local anchor points for predicting offsets between them and joints lack interaction among each other. This leads to the fact that, joints' global articulated clues cannot be well captured to resist occlusion. Secondly, the anchor points within the certain spatial range share the same single-scale local convolution feature, which essentially limits the discrimination capacity on confusing visual patterns towards the interacting hands. Last, anchor points locate within 2D plane, which is not optimal for alleviating the ill-posed 2D to 3D lifting problem with single RGB image. To address these, we propose to \emph{extend A2J under Transformer's non-local encoding-decoding framework to build A2J-Transformer, with anchor point-wise adaptive multi-scale feature learning and 3D anchor point setup}.

Particularly, the anchor point within A2J is evolved as the learnable query under Transformer framework. Each query will predict its position offsets to all the joints of the 2 hands. Joint's position is finally estimated via fusing the prediction results from all queries in a linear weighting way. That is to say, joint's position is determined by all the queries located over the whole image of global spatial perspective. Meanwhile, the setting query number is flexible, which is not strictly constrained by joint number as in~\cite{hampali2022keypoint}. Thanks to Transformer's non-local self-attention mechanism~\cite{vaswani2017attention}, during feature encoding stage the queries can interact with each other to capture joints' global articulated clues, which is essentially beneficial for resisting self-occlusion and inter-occlusion. Concerning the specific query, adaptive local feature learning will be conducted to extract query-wise multi-scale convolutional feature based Resnet-50~\cite{he2016deep}. Compared with A2J's feature sharing strategy among the neighboring anchor points, our proposition can essentially facilitate query's pattern fitting capacity both for accurate joint localization and joint's hand identity verification. In summary, each query will be of strong local-global spatial awareness ability to better fit interacting hand appearance pattern. Meanwhile to facilitate RGB-based 2D to 3D hand pose lifting problem, the queries will be set within the 3D space instead of 2D counterpart as in A2J~\cite{xiong2019a2j}. In this way, each query can directly predict its 3D position offset between the joints, which cannot be acquired by A2J. Overall, A2J-Transformer's main research idea is shown in Fig.~\ref{fig:figure1}.

Compared with the most recently proposed model-free method~\cite{hampali2022keypoint} that also addresses 3D interacting hand pose estimation using Transformer, our proposition still takes some essential advantages. First, joint-like keypoint detection is not required. Secondly, query number is not strictly constrained to be equal to joint number to facilitate pattern fitting capacity. Thirdly, our query locates within 3D space instead of 2D counterpart.

The experiments on the challenging Interhand 2.6M~\cite{moon2020interhand2} dataset verify that, our approach can achieve the state-of-the-art model-free performance (3.38mm MPJPE advancement in 2-hand case) for 3D interacting hand pose estimation from a single RGB image. And, it significantly outperforms A2J by large margins (i.e., over 5mm on MPJPE). In addition, experiments on HANDS2017 dataset~\cite{yuan20172017} demonstrate that A2J-Transformer can also be applied to depth domain with promising performance.

Overall, the main contributions of this paper include:

$\bullet$ For the first time, we extend A2J from depth domain to RGB domain to address 3D interacting hand pose estimation from a single RGB image with promising performance;

$\bullet$ A2J's anchor point is evolved with Transformer's non-local self-attention mechanism with adaptive local feature learning, to make it be aware of joints’ local fine details and global articulated context simultaneously;

$\bullet$ Anchor point is proposed to locate within 3D space to facilitate ill-posed 2D to 3D hand pose lifting problem based on monocular RGB information.

\begin{figure*}[t]
\centering
\includegraphics[width=0.84\linewidth]{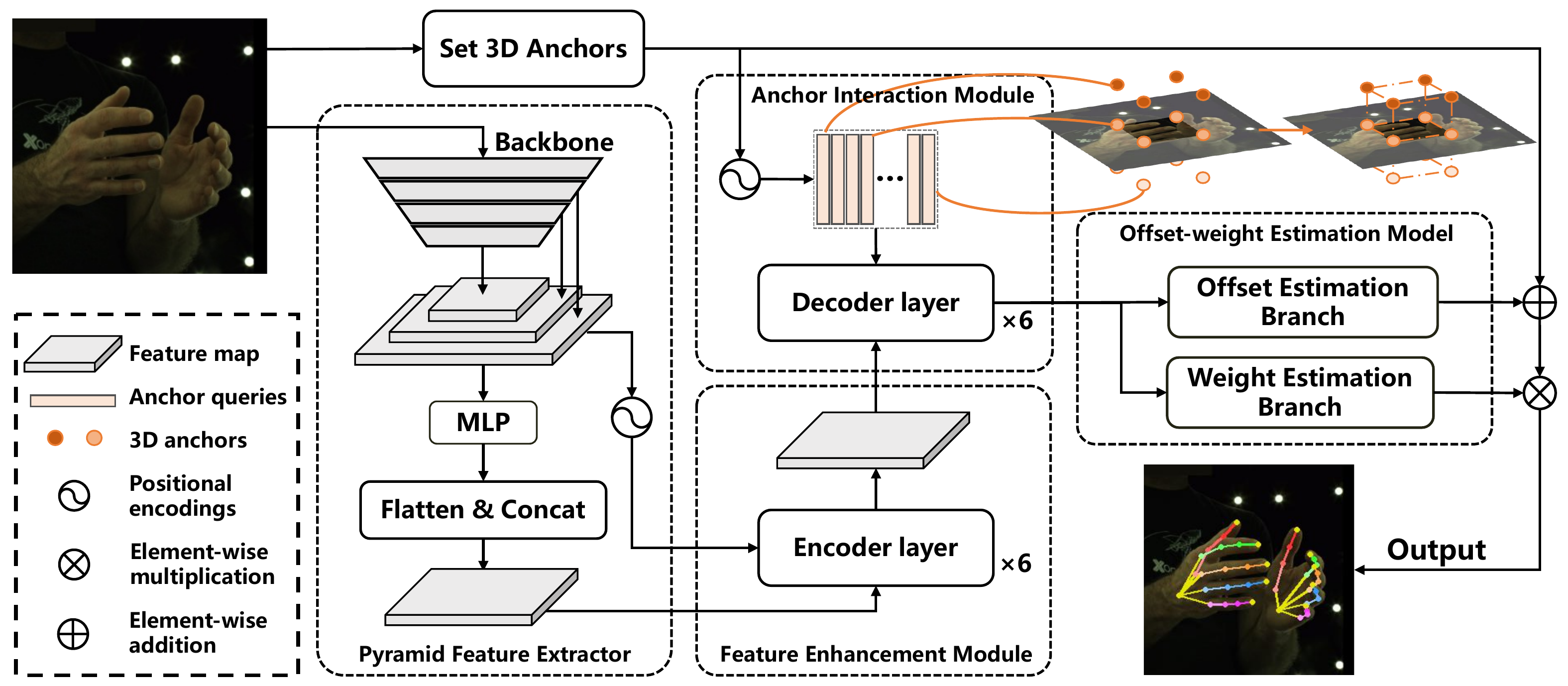}
\caption{The main technical pipeline of A2J-Transformer. 
A2J-Transformer consists of 3 main models: pyramid feature extractor, anchor refinement model (containing feature enhancement module and anchor interaction module) and anchor offset-weight estimation model. The anchor interaction module aims to establish the connection (orange line) between anchors (orange dots).}
\label{fig:pipline}
\vspace{-12pt}
\end{figure*}

%-------------------------------------------------------------------------------------
%The most similar work to us is~\cite{hampali2022keypoint} that also addresses 3D interacting hand pose estimation using Transformer. However, our approach essentially takes 3 main advantages.   

\section{Related Works} 
\label{Related Works}
Many methods have been proposed for 3D hand pose estimation from either RGB images or depth maps.
% Here we mainly discuss works focus on RGB images.
At the same time, these methods can also be divided into single hand pose estimation and interacting hand pose estimation methods based on the number of input hands.
Here we categorize all these methods into model-based and model-free groups, and mainly analysis works that estimate interacting 3D hand pose from RGB images.
Besides, we discuss the usage of Transformer architectures in 3D hand pose estimation field as they are highly relevant to our work.

\textbf{Model-based approach.}
Considering that model-based methods~\cite{oikonomidis2012tracking,ballan2012motion,tzionas2016capturing,wang2020rgb2hands,romero2022embodied,smith2020constraining,moon2020interhand2,zhang2021interacting,li2022interacting} can provide strong prior knowledge, model-based 3D hand pose estimation methods could achieve relatively better results by fitting hand models.
Early methods~\cite{oikonomidis2012tracking,ballan2012motion,tzionas2016capturing} for model-based 3D hand pose estimation used complex optimization methods to fit their proposed parameter models.
However, due to the lack of a unified model paradigm, the development of model-based methods was somewhat limited at that time.
After the introduction of the compatible 3D hand model MANO~\cite{romero2022embodied}, subsequent model-based methods~\cite{wang2020rgb2hands,smith2020constraining,zhang2021interacting,li2022interacting} are mostly based on it while using CNN or GCN modules.
Due to the presence of a sufficiently strong prior model, model-based methods generally have good performance and are more stable than model-free methods.
However, these approaches usually lose tracking when there are strong hand interactions and occlusions.
At the same time, modeling the hands of different people is needed~\cite{han2022umetrack} in practical usage, which to a certain extent reduces the generalization ability of the model.
% in practical applications, the model-based approaches need to model the hands of different people~\cite{han2022umetrack}, which to a certain extent reduces the generalization ability of the model.
Therefore, we turn our attention to the model-free approach, which needs no prior information and has more flexibility.

\textbf{Model-free approach.}
Model-free approaches~\cite{xiong2019a2j,cai2018weakly,moon2018v2v,iqbal2018dual,moon2020interhand2,cheng2021handfoldingnet,fan2021learning,lin2021two,kim2021end,meng20223d} have been developed for a long time.
In particular, the task of single hand pose estimation based on depth maps have been available for very mature methods~\cite{xiong2019a2j,cai2018weakly,moon2018v2v,iqbal2018dual}.
However, their extensions to two hands and RGB domain are non-trivial due to the severe occlusion and similar appearance between hand joints.
% Xiong \etal~\cite{xiong2019a2j} propose a CNN based hand pose estimation network which has a strong ability in capturing local features and achieves the state-of-the-art performance with fast inference speed.
% However, this architecture is designed for single hand estimation form depth image.
% The extension to two hands and RGB domain is non-trivial.
% Due to the lack of dataset, Cai \etal~\cite{cai2018weakly} propose a weakly-supervised method which uses depth images for training and only RGB images for testing.
% Iqbal \etal~\cite{iqbal2018dual} propose a dual-source approach that combines 2D pose estimation with efficient 3D pose retrieval to address this problem.
After Moon \etal~\cite{moon2020interhand2} propose the \textit{InterHand2.6M} dataset, model-free approaches~\cite{cheng2021handfoldingnet,fan2021learning,lin2021two,kim2021end,deng2022recurrent,meng20223d} for interacting hand pose estimation has made great progress.
For be better resistant to occlusion, some research~\cite{fan2021learning,lin2021two,meng20223d} tend to separate the interacting hands and estimate the two hands separately, some methods~\cite{cheng2021handfoldingnet,deng2022recurrent} perform dense modeling by using point cloud networks.
However, the prediction of details of interacting hands by these methods depend heavily on the quality of the segmentation results or the point cloud generations.
Some methods~\cite{moon2020interhand2,hampali2022keypoint} obtain the coordinate of hand joints by directly regressing the heatmap, which could be intuitive and flexible.
However, the current methods are not ideal for local detail feature extraction and still have performance shortcomings for 3D interacting hand pose estimation.
% To deal with self-similarity of interacting hands, Fan \etal~\cite{fan2021learning} leverage the per-pixel probabilities to process the input imagery into a per-pixel semantic part segmentation mask and a visual feature volume.
% Lin \etal~\cite{lin2021two} also use segmentation approaches to locate the hand despite occlusion and noisy background and propose a new RGB interacting hand dataset.
% Kim \etal~\cite{kim2021end} achieve end-to-end two hand pose estimation which employed a GAN-type discriminator of interacting hand pose that helps avoid physically implausible configurations.
% Recently, Meng \etal~\cite{meng20223d} decompose the interacting hand pose estimation task and estimate the pose of each hand separately using hand de-occlusion and removal.
% These works try to segment the two hands from images and estimate them separately, which could improve the occlusion of interacting hands and the problem of similar patterns.
The proposed method of Hampali \etal~\cite{hampali2022keypoint} is similar to ours, which directly regress the keypoints of two hands, but there is still a problem of poor prediction effect when having strong occlusions. 
In contrast, by regarding densely distributed anchor points as local regressioners and establishing interactions between them, our proposed A2J-Transformer can not only extract local detailed hand poses, but also obtain global articulated hand joints' information.

\textbf{Transformer in hand pose estimation.}
With the rise of the self-attention mechanism and the proposal of the transformer model~\cite{vaswani2017attention}, more and more visual fields promote their development by introducing the transformer model, like image classification, object detection, 3D mesh reconstruction and so on~\cite{khan2022transformers}.
Since the Transformer model has a strong ability in capturing non-local features which is surely helpful for the hand pose estimation field, there has been many works~\cite{lin2021end,lin2021mesh,huang2020hand} to introduce Transformer into this area.
% Lin \etal~\cite{lin2021end} propose to reconstruct 3D human and hand pose by multiple transformer encoder layers.
% Lin \etal~\cite{lin2021mesh} expand their work by combining Transformer encoder with graph convolutions.
% Huang \etal~\cite{huang2020hand} estimate 3D hand pose from point cloud and use Transformer encoder and decoder layers to provide equivalent dependencies among hand joints.
However, these architectures are all designed for single hand pose estimation.
Recent methods for interacting hand pose estimation have achieved good results, but still suffer from performance shortcomings~\cite{hampali2022keypoint} or model limitations in flexibility~\cite{li2022interacting}.
% Hampali \etal~\cite{hampali2022keypoint} propose an interacting hand and object pose estimation architecture from single RGB image by using Transformer encoder and decoder layer to associate the joint keypoints with the correct position.
% As mentioned above, this method still has certain problems due to the inability to recover local details.

% After DETR~\cite{carion2020end} introduced Transformer into object detection, many related works~\cite{meng2021conditional,zhu2020deformable,liu2022dab} have improved and perfected DETR.
% Liu \etal~\cite{liu2022dab} gives a deeper understanding of the role of queries in DETR.
% They present a novel query formulation by directly setting learning anchors as queries and perform dynamic anchor update layer-by-layer.
% Inspired by them, our work takes the understanding %of them 
% and extends them to the field of interacting hand pose estimation.
% We explicitly set the queries in Transformer decoders as the 3D coordinates of the anchor points, and regard the anchor points as regressors to estimate the hand joints and fuse them by weights obtained from learning.

Accordingly, our A2J-Transformer belongs to model-free region and introduced the Transformer module.
Different from previous works, we integrates A2J and Transformer into an uniform model (i.e., A2J-Transformer) with end-to-end learning capacity, to reveal our key theoretical insight on addressing 3D interacting hand pose estimation (IHPE) task via concerning local and global visual context jointly. Meanwhile, 2D anchor point within A2J is evolved to 3D version adaptive to A2J-Transformer, to alleviate ill-posed 2D to 3D hand pose lifting problem using monocular RGB image. These propositions technically sound with promising performance and concern 3D IHPE's specific characteristics deeply.

%----------------------------------------------------------------------------------------------
%section 3
\section{A2J-Transformer: Anchor-to-Joint Transformer Network} 
\label{A2J-Transformer: Anchor-to-Joint Transformer Network}

\renewcommand\arraystretch{1.2}

As shown in Fig.~\ref{fig:pipline}, A2J-Transformer consists of 3 main models: pyramid feature extractor, anchor refinement model and anchor offset-weight estimation model.

\subsection{Pyramid feature extractor}
Since multi-scale features can obtain both global information of the input image and retain enough detailed information, feature pyramids are well suitable for the task of interacting hand pose estimation.
Therefore, ResNet-50~\cite{he2016deep} is used as backbone network to extract the pyramid features from input RGB images.
In particular, we get the pyramid features by using the output layer 2-4 with 8,16,32$\times$ down sample rates on in-plane size.
Meanwhile, 3 convolution layers are used for generating inputs of transformer model of each feature, and 1 convolution layer is used additionally for extracting the last feature layer to maintain more spatial information.
% and generate the input features for next transformer model.
% For a cropped 256$\times$256 RGB image, we used ResNet-50 pre-trained on ImageNet as the backbone network.
% In particular, we get the pyramid features by using the output layer 2-4 with 8,16,32$\times$ downsampling on in-plane size, whose channels are 512,1024,2048 relatively.
% Meanwhile, 3 convolution layers are used for generating inputs of transformer model of each feature, and 1 convolution layer is used additionally to extract the last feature layer to maintain more spatial information.
Finally, these 4 feature maps are sent to the next anchor refinement model.

\subsection{Anchor refinement model}
Anchor refinement model aims to simultaneously focus on the non-local articulated and the local fine-grained features.
It contains feature enhancement module and anchor interaction module, which can 
 enhance image features and establish the interactions between anchors respectively.

\begin{figure}
\centering
\includegraphics[width=0.85\linewidth]{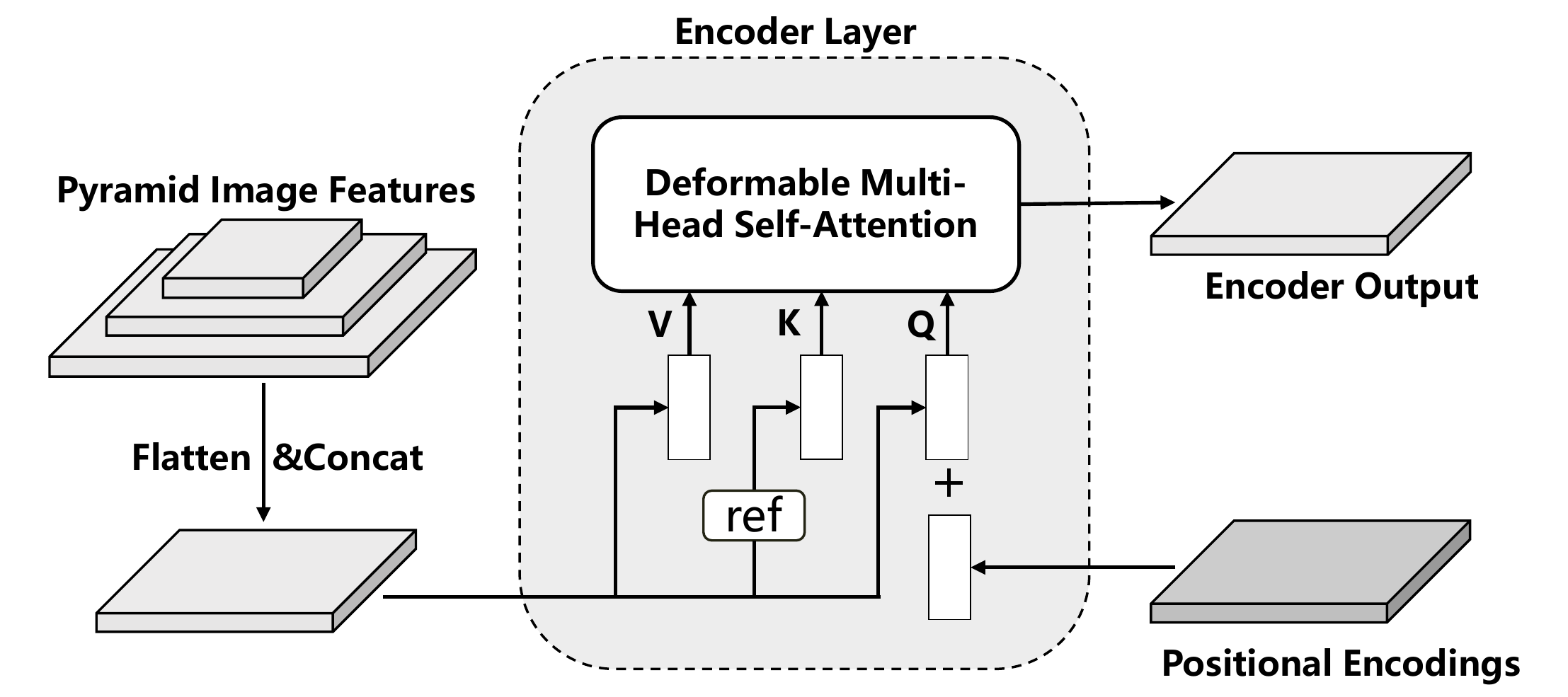}
\caption{The first encoder layer of feature enhancement module.}
\label{fig:encoder}
\vspace{-12pt}
\end{figure}

\vspace{-12pt}
\subsubsection{Feature enhancement module}
\vspace{-3pt}
\
\indent
Since multi-scale features are useful for capturing global clues and recovering local details, we integrated the self-attention module~\cite{liu2022dab} to enhance multi-scale features.
So we refer to this module as feature enhancement module, which consists of six encoder layers.
The first encoder layer of this module is shown in Fig.~\ref{fig:encoder}, and the input features of the rest encoder layers are the outputs of the previous layers.
All dimensions of input and output features are 256.

Technically, for the input feature pyramid, convolution layer and group normalization layer\wcl{~\cite{wu2018group}} are firstly used to process them to \wcl{a} %the 
same in-plane size.
After flatten and concatenation, the generated features $F$ are added to the positional encodings $P_{xy}$ :
\vspace{-3pt}
\begin{equation}
\label{eq:encoderpe}
P_{xy} = \rm{PE}(x,y)\wcl{,}
\end{equation}
where PE means positional encoding to generate sinusoidal embeddings from float numbers~\cite{liu2022dab}, and $x$,$y$ represent the in-plane positions of the feature $F$.
Besides, we replace the self-attention module with multi-scale deformable attention module (MSDAM)~\cite{zhu2020deformable} to mitigate issues of slow convergence and limited feature resolution.

For self-attention module, the queries $Q$,  keys $K$ and values $V$ have the same content item $F$, and the queries contains an extra position item $P_{xy}$:
\vspace{-3pt}
\begin{equation}
\label{eq:encoderQKV}
Q = F + P_{xy},\quad  K = ref (F), \quad  V = F,
\end{equation}
where $ref(\cdot)$ means sample reference keys following  ~\cite{zhu2020deformable}.
Then, $Q,K,V$ are sent to the MSDAM to get the enhanced features for next encoder layers.

Finally, global-aware features are generated after 6 encoder layers and sent to the anchor interaction module.

\begin{figure}
\centering
\includegraphics[width=0.8\linewidth]{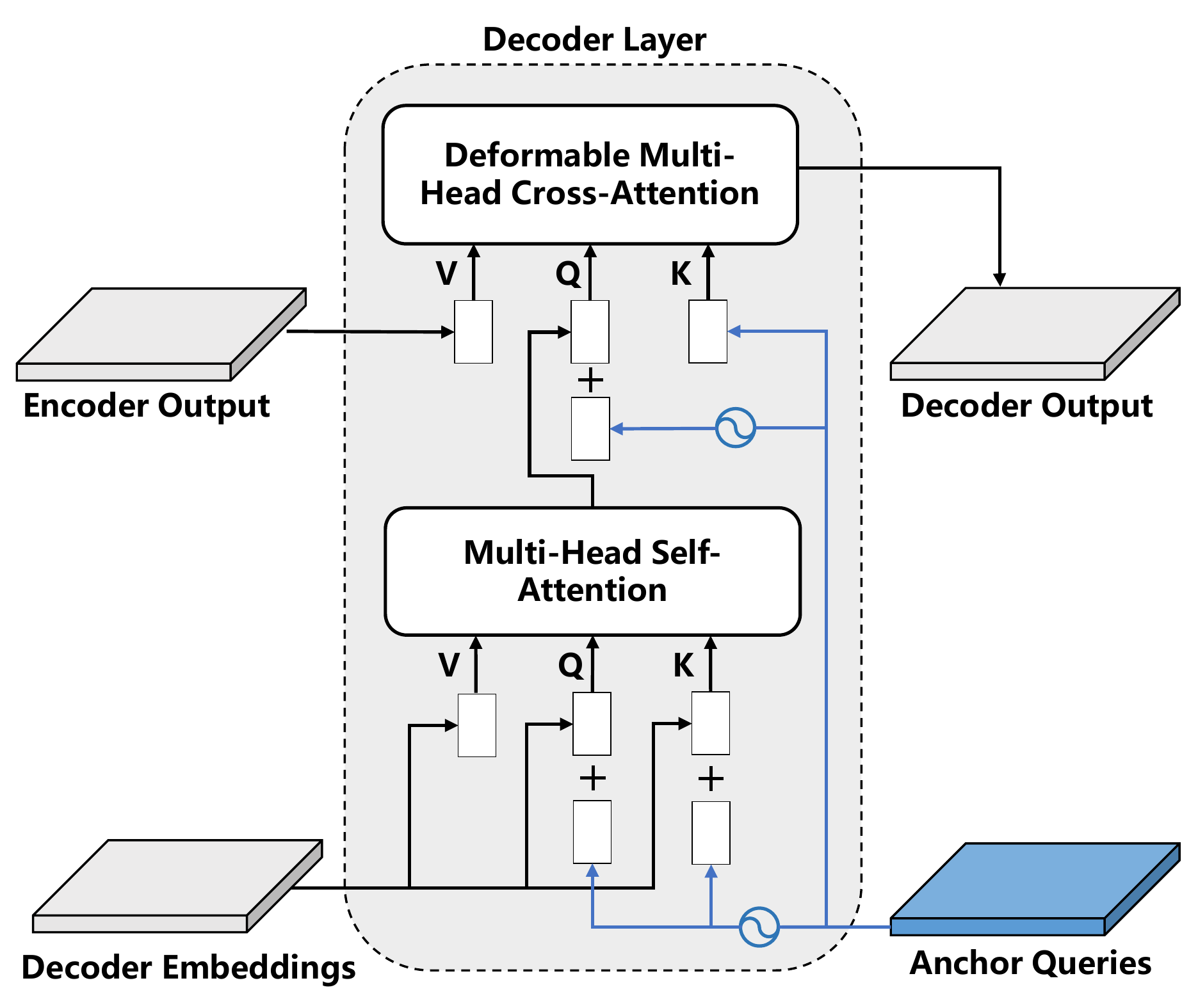}
\caption{One decoder layer of anchor interaction module.}
\label{fig:decoder}
\vspace{-12pt}
\end{figure}

\vspace{-8pt}
\subsubsection{Anchor interaction module}
\vspace{-12pt}
\ 
\indent

In A2J-Transformer, a uniform distribution of 3D anchor points are  densely set up to perform direct estimation of hand joints through these 3D anchor points.
In other words, these 3D anchor points play the role of local coordinate regressors.
More details on the implementation of 3D anchor settings are described in Sec.~\ref{Anchor offset-weight estimation model}.
Estimating hand pose through local anchor points has two advantages.
First, the setting of dense local 3D anchor points can effectively capture the refined local details from images, which has a good effect for estimating the detail information of strong interacting hands.
Second,cross-attention module can establish interaction between local anchor points to capture global clues, which is beneficial for handling occlusion.

Based on this, anchor interaction module containing 6 decoder layers are designed to link individual anchor points, making global information available for each anchor point.
One decoder layer is shown in Fig.~\ref{fig:decoder} and for the first decoder layer, the Decoder Embeddings will be replaced by the Encoder Output.
All dimensions of input and output features are 256.

\begin{table}
\scriptsize
\centering
\begin{tabular}{cc}
\toprule
Symbol & Definition \\
\hline\hline
$J$ \& $j$ & Joint set and joint.$j\in J$.\\
$A$ \& $a$ & Anchor point set and anchor point.$a\in A$.\\
$T^i_j$ & In-plane coordinate of joint $j$.\\
$T^d_j$ & Depth coordinate of joint $j$.\\
$C^i(a)$ & In-plane coordinate of anchor point $a$.\\
$C^d(a)$ & Depth coordinate of anchor point $a$.\\
$W_j(a)$ & Weight of anchor $a$ towards joint $j$. \\
$O^i_j(a)$ & Predicted in-plane offset towards joint $j$ from anchor point $a$.\\
$O^d_j(a)$ & Predicted depth offset towards joint $j$ from anchor point $a$.\\
\bottomrule
\end{tabular}
\caption{Symbol definition within A2J-Transformer.}
\label{tab:symbol}
\vspace{-12pt}
\end{table}

The symbols within A2J-Transformer are defined in Table~\ref{tab:symbol} for better explaining.
Different from previous Transformer-based works, we take the understanding of DAB-DETR~\cite{liu2022dab} and explicitly set the coordinates of each anchor $a$ to the queries, which we call Anchor Queries.
We denote $a_q=\left( {x_q},{y_q},{d_q}\right)$ as the $q$-th anchor query, while $x_q,y_q,d_q \in \mathbb{R}$ denotes the coordinate of $a$ in in-plane and depth.
For $a_q$, the spatial encodings $P_q$ is generated by:
\vspace{-3pt}
\begin{equation}
\label{eq:pq}
{P_q} = {\rm{MLP}}({\rm{PE}}({a_q})),
\end{equation}
where parameters of MLP are shared across all layers. 

For self-attention module, settings of queries, keys and values of decoder layers are similar to the setting in feature enhancement module:
\vspace{-3pt}
\begin{equation}
\label{eq:decoderQKV1}
Q = D + P_q,\quad  K = D + P_q,\quad  V = D,
\end{equation}
where $D$ denotes the decoder embeddings.

In cross-attention module, we add the positional query embeddings $P_q$ to the output of self-attention module $D$ to get the context aware anchor informative queries $Q$. 
Besides, anchor queries are directly set to the reference points $K$, and $V$ is the encoder output $E$:
\vspace{-3pt}
\begin{equation}
\label{eq:decoderQKV2}
Q = D + P_q, \quad   K = a_q, \quad   V = E,
\end{equation}
and MSDAM is applied for calculating cross-attention.

\subsection{Anchor offset-weight estimation model}
\label{Anchor offset-weight estimation model}
As described in Section 3.2, when each anchor point is linked to each other through the Transformer module, they have both the ability to recover local details and perceive global information.
To get final output, anchor offset-weight estimation model is used to estimate the 3D offsets and weights of each anchor with respect to each hand joints. 
That is, each anchor acts as a local estimator.
The offsets and weights are estimated separately for all hand joints.
Finally, we fuse the estimation results of all anchors in a weighted summation way to get the final result of joints.

\begin{figure}[t]
\centering
\includegraphics[width=0.7\linewidth]{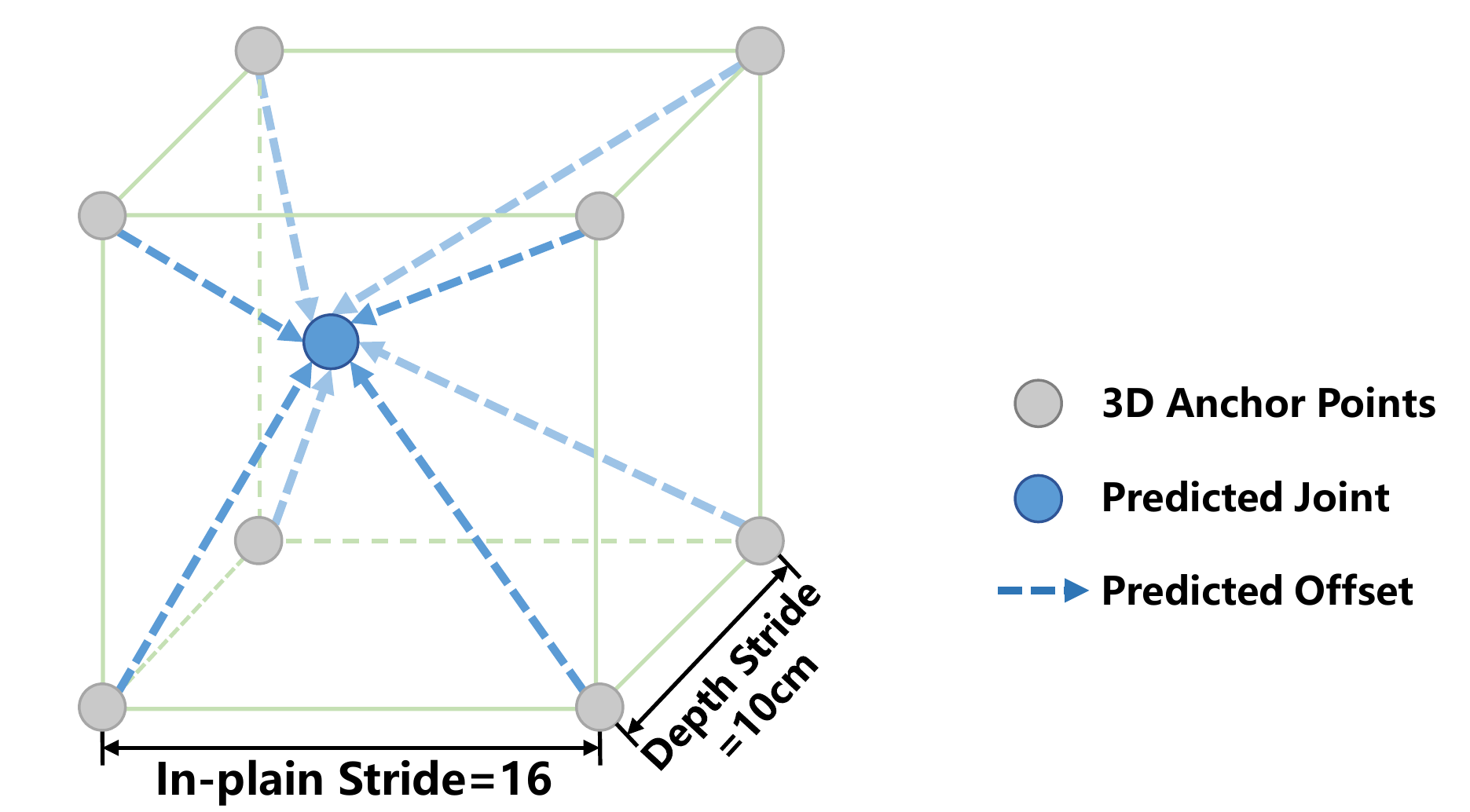}
\vspace{-4pt}
\caption{3D anchors in A2J-Transformer. Joints will be estimated from anchors and offsets.}
\label{fig:anchor}
\vspace{-12pt}
\end{figure}

The 3D anchor structure is shown in Fig.~\ref{fig:anchor}.
The in-plane coordinates of 3D anchors are densely distributed on the input RGB image with an in-plane stride $S_t=16$.
This could ensure that for each pixel in the extracted feature maps, there can be at least one anchor point corresponding to it while reducing the model size.
On this basis, we extend the depth value number of the anchor points.
In addition to the original 0 depth value, two depth values are set at the position of $\pm$100 mm under the world coordinate, centered on the root joint of each hand.
This is due to the data processing procedure within baseline. 
That is, hand joints outside the range of ±200mm from the root of the
hand are considered as invalid joints.
Therefore, for the input image size 256$\times$256, there are 16$\times$16$\times$3 total anchors.
This setting method extends the anchor point to the 3D space, so as to better fit the depth coordinates of the predicted joints.

Essentially, anchor points are local regressors used to estimate each joint relative to itself.
As shown in Fig.~\ref{fig:pipline}, each anchor point returns a 3D coordinate offset from itself to all joints in \textbf{offset estimation branch}.
Since different anchor points focus on different feature ranges, the contribution to each anchor point will also be different.
So we predict the weight of each anchor point by \textbf{weight estimation branch}.
Therefore, by these two branches, the coordinates of each joint can be calculated as the weighted sum of the prediction results of all anchor points' coordinates.

Technically, to get the offsets ${O^i_j(a)}$ , ${O^d_j(a)}$ and the anchor weights $W_j(a)$, 2 MLP layers are added on the outputs of anchor interaction model.
3D offsets from each anchor point to each joint ${O_j(a)}$ are regressed by \wcl{1 }%one 
MLP layer and then divided into ${O^i_j(a)}$ , ${O^d_j(a)}$.
Another MLP layer regresses each anchor weight $W_j(a)$ for each joint.
\wcl{Finally, the 3D coordinates of predicted joint $j$ can be expressed as:}
\vspace{-3pt}
\begin{equation}
\label{eq:A2J}
\begin{aligned}
\left\{ \begin{array}{l}
{{\hat T}^i_j} = \sum\limits_{a \in A} {{{\tilde W}_j}\left(a\right)\left({C^i\left(a\right)} + {O^i_j}\left(a\right)\right)} \\
{{\hat T}^d_j} = \sum\limits_{a \in A} {{{\tilde W}_j}\left(a\right)\left({C^d\left(a\right)} + {O^d_j}\left(a\right)\right)}
\end{array} \right.,
\end{aligned}
\end{equation}
where ${{\hat T}^i_j}$ and ${{\hat T}^d_j}$ indicate the estimated in-plane and depth coordinate of target joint $j$, $C^i_a$ and $C^d_a$ denote the in-plane and depth coordinates of an anchor point $a$.
$\tilde W_j(a)$ is the normalized weight from anchor point $a$ towards joint $j$, which could be calculated by soft-max:
\vspace{-3pt}
 \begin{equation}
\label{eq:softmax}
\begin{aligned}
 \tilde W_j(a) = \frac{{{e^{{W_j}(a)}}}}{{\sum\limits_{a \in A} {{e^{{W_j}(a)}}} }}.
 \end{aligned}
\end{equation}

In this way, the estimated hand joints will adaptively select those anchor points with greater contributions to itself and give them large weights.
% We supervise the joint coordinates and the anchor weights through joint estimation loss and anchor point surrounding loss.
Finally, the joint coordinates and the anchor weights are supervised through joint estimation loss and anchor point surrounding loss.

\subsection{Loss functions}
For training our performed A2J-Transformer model, we utilize two loss functions: (1) joint estimation loss, (2) anchor surrounding loss following~\cite{xiong2019a2j}.

\textbf{Joint estimation loss}.
After getting the estimated 3D joint coordinates, we use the joint estimation loss to supervise the final output, which is formulated as:
\vspace{-2pt}
 \begin{equation}
\label{eq:loss1}
\begin{aligned}
{{loss}_1} = \alpha \sum\limits_{j \in J} {{L_{\tau_1}}(\hat{T}_j^i - T_j^i)}  + \sum\limits_{j \in J} {{L_{\tau_2}}(\hat{T}_j^d - T_j^d)},
 \end{aligned}
\end{equation}
where $\hat{T}_j^i$ and $\hat{T}_j^d$ denotes the estimated in-plane coordinate and depth coordinate of joint $j$ from Eq.\ref{eq:A2J}, and $T_j^i$ and $T_j^d$ are the given in-plane and depth GT coordinates of joint $j$; 
parameter $\alpha$ defaults to 0.5 to balance the loss between in-plane and depth offset estimation task.
$L_{\tau}(\cdot)$ is the $smooth_{L1}$ like loss function~\cite{ren2015faster} given by:
\vspace{-3pt}
\begin{equation}
\label{eq:smoothL1}
\begin{aligned}
{L_\tau(x) } = \left\{ \begin{array}{l}
\frac{1}{{2\tau }}{x^2},{\qquad \textnormal{for} \ }\left| x \right| < \tau ,\\
\left| x \right| - \frac{\tau }{2},{\quad }\textnormal{otherwise}.
\end{array} \right.
\end{aligned}
\end{equation}
$\tau_1$, $\tau_2$ are set to $1$, $3$ for better smoothing the depth value.

\textbf{Anchor surrounding loss}.
To lead the informative anchor points locate around the hand joints, thus facilitating the generalization ability of our model, we define the anchor surrounding loss by: 
\vspace{-3pt}
\begin{equation}
\label{eq:loss2}
\centering
\begin{aligned}
\begin{array}{l}
los{s_2} = \sum\limits_{j \in J} {{L_{{\tau _1}}}(\sum\limits_{a \in A} {{{\tilde W}_j}(a){C^i}(a) - T_j^i)} } \\
 + \sum\limits_{j \in J} {{L_{{\tau _2}}}(\sum\limits_{a \in A} {{{\tilde W}_j}(a){C^d}(a) - T_j^d)} },
\end{array}
\end{aligned}
\end{equation}
where $\tau_1$ and $\tau_2$ are also set to 1 and 3.

Finally, the total loss function is formulated as:
\vspace{-3pt}
\begin{equation}
\label{eq:loss}
loss = {\lambda _1}{{loss}_1} + {\lambda _2}{{loss}_2}.
\end{equation}
where $\lambda _1$ and $\lambda _2$ are set to 3 and 1 to balance two losses.

%---------------------------------------------------------------------------------------
\section{Experiments}

\subsection{Experimental setting}
\subsubsection{Datasets}
\vspace{-4pt}
\noindent \textbf{InterHand2.6M dataset}~\cite{moon2020interhand2}.
InterHand2.6M is a representative two-hand RGB image dataset with challenging hand interacting scenarios.
It contains 1.36M train images and 849K test images.
The ground-truth contains semi-automatically annotated 3D coordinates of 42 hand joints.
For fair comparison, we choose all test frames for result evaluation following InterNet~\cite{moon2020interhand2}.

\noindent \textbf{RHP dataset}~\cite{zimmermann2017learning}.
RHP is a synthesized dataset contains two isolated hand data.
41K training and 2.7K testing samples are contained.
Since the background of this dataset is an outdoor scene, we use this dataset to approximate the generalization ability of our model on in-the-wild conditions.
We also follow InterNet for fair comparison.

\noindent \textbf{NYU dataset}~\cite{tompson2014real}.
NYU is a single-hand depth image dataset which has 72K training images and 8.2K testing images with 3D annotation on 36 hand joints.
Following~\cite{guo2017region,moon2018v2v,chen2020pose,xiong2019a2j}, we pick 14 of the 36 joints for evaluation.

\noindent \textbf{HANDS 2017 dataset}~\cite{yuan20172017}.
HANDS 2017 is a single-hand depth image dataset which has 957K training images and 295K testing images combined from BigHand2.2M~\cite{yuan2017bighand2} and First-Person Hand Action~\cite{yuan20172017}.
The ground-truth contains 3D coordinate of 21 hand joints.

\vspace{-12pt}
\subsubsection{Evaluation metrics}
\vspace{-4pt}
\ 
\indent
The \textbf{Mean Per Join Position Error (MPJPE)} is used for evaluation on InterHand2.6M~\cite{moon2020interhand2}.
It is defined as a Euclidean distance (mm) between predicted and ground-truth 3D joint locations.
Following~\cite{moon2020interhand2}, this metric is used after root joint alignment for each left and right hand separately.
For RHP dataset, \textbf{end point error (EPE)} is used, which is defined as a mean Euclidean distance (mm) between the predicted and ground-truth 3D hand pose after root joint alignment.
For the two depth image dataset, the \textbf{average 3D distance error} is used following~\cite{moon2018v2v,xiong2019a2j}.
Besides, \textbf{FPS} is used to evaluate the inference speed, and all models are tested on single NVIDIA RTX 2080ti GPU during inference.

\vspace{-12pt}
\subsubsection{Implementation details}
\vspace{-4pt}
\ 
\indent
A2J-Transformer is implemented using PyTorch.
For InterHand2.6M and RHP dataset, we directly crop the RGB images and resize them to 256$\times$256 resolution. 
The data augmentations are exactly the same as InterNet~\cite{moon2020interhand2}.
For NYU and HANDS 2017 dataset, we follow~\cite{moon2018v2v} to crop and resize the depth image to 176$\times$176.
We train our model using the Adam optimizer~\cite{kingma2014adam}.
The learning rate is set to $1\times{10^{-4}}$ with a weight decay of $1\times{10^{-4}}$ in all cases.
There are totally 42 epochs for InterHand2.6M, RHP and NYU dataset and 17 epochs for HANDS 2017 dataset.

\begin{table}
\vspace{-4pt}
\scriptsize
\centering
\begin{tabular}{lccccc}%四个c代表有四列且内容居中
\toprule%第一道横线
\multirow{2}{*}{Methods} & \multicolumn{3}{c}{MPJPE (mm)} & FPS & Model\\
                           % & \multicolumn{3}{c}{MPJPE (mm)} & FPS & Model\\
\cline{2-4}
                         &  Single  &  Two  &  All  & (s)  & Size(M) \\
\midrule%第二道横线 
\multicolumn{6}{c}{\textbf{Model-based}} \\
\midrule
% \multirow{2}{*}{\makecell[c]{Model \\ based}}
Zhang \etal~\cite{zhang2021interacting} & -     & 13.48 & -     & 17.02 &  143 \\
Meng \etal~\cite{meng20223d}            & 8.51  & 13.12 & 10.97 & 15.47 &  55  \\
Li \etal~\cite{li2022interacting}       & -     & 8.79  & -     & 18.05 &  39  \\
\midrule
\multicolumn{6}{c}{\textbf{Model-free}} \\
\midrule
% \multirow{4}{*}{\makecell[c]{Model \\ free}}
Moon \etal~\cite{moon2020interhand2} & 12.16 & 16.02 & 14.22 & \textbf{107.08} & 47 \\
Kim  \etal~\cite{kim2021end}            & -     & -     & 12.08 & - & -   \\
Fan \etal~\cite{fan2021learning}        & 11.32 & 15.57 & -     & - & -  \\
Hampali \etal~\cite{hampali2022keypoint}& 10.99 & 14.34 & 12.78 & 19.66 & 48 \\
% \midrule%第三道横线 
Ours  & \textbf{8.10}  & \textbf{10.96} & \textbf{9.63}  & 25.65 & \textbf{42} \\
\bottomrule%第四道横线
\end{tabular}
\vspace{-4pt}
\caption{Comparison with state-of-the-art model-based and model-free methods on InterHand2.6M~\cite{moon2020interhand2}.
MPJPE, FPS and model size are reported.}
\vspace{-4pt}
\label{tab:Interhand}
\end{table}

\subsection{Results}
\textbf{InterHand2.6M dataset}:
Comparison with the state-of-the-art methods on InterHand2.6M is listed in Table~\ref{tab:Interhand}.
It can be observed that:

$\bullet$ In general, A2J-Transformer outperforms other model-free methods by a large margin testing under all scenarios. 
This proves that our method has a significant advancement in extracting effective information from interacting hands.
Compared with model-based methods, A2J-Transformer has a comparable result with the state-of-the-art method without using any hand prior information.
Besides, A2J-Transformer has fairly fast inference speed just behind baseline~\cite{moon2020interhand2} and the smallest model size.
In conclusion, our model achieves the best overall performance in terms of performance, running speed and model size.

$\bullet$ Specifically, compared with baseline~\cite{moon2020interhand2}, A2J-Transformer could get an improvement of 4.06, 5.06 and 4.59mm under three scenarios.
Compared with the SOTA model-free method~\cite{hampali2022keypoint}, the improvement of our method is 2.89, 3.38 and 3.15mm.
% It can be seen that A2J-Transformer can achieve greater performance improvement under the interacting hand scenario than the single hand scenario, which proves that our method is much stronger than other methods in extracting details and global features.
Compared with the SOTA model-based method~\cite{li2022interacting}, our method could receive a comparable performance under two hands scenario without requiring any hand prior, which makes our model more flexible.

$\bullet$ For the running speed, A2J-Transformer has a fast inference speed with 25 FPS, surpassing all methods except baseline.
Besides, A2J-Transformer also has the smallest model size with only 42M parameters.
These characteristics brings our model great convenience for the future expansion and real-time 3D hand pose estimation.

\begin{table}
\scriptsize
\centering
\begin{tabular}{lccc}%四个c代表有四列且内容居中
\toprule%第一道横线
Methods  &  GT S  &  GT H  &  EPE  \\
\midrule%第二道横线 
Zimm. \etal~\cite{zimmermann2017learning}   & \checkmark & \checkmark & 30.42  \\
chen \etal~\cite{chen2018generating}        & \checkmark & \checkmark & 24.20  \\
Yang \etal~\cite{yang2019disentangling}     & \checkmark & \checkmark & 19.95  \\
Spurr \etal~\cite{spurr2018cross}           & \checkmark & \checkmark & 19.73  \\
Spurr \etal~\cite{spurr2018cross}         & \XSolidBrush & \XSolidBrush & 19.73\\
Moon \etal~\cite{moon2020interhand2}      & \XSolidBrush & \XSolidBrush & 20.89\\
A2J-Transformer(Ours)   & \XSolidBrush & \XSolidBrush &  \textbf{17.75}  \\
\bottomrule%第四道横线
\end{tabular}
\vspace{-4pt}
\caption{EPE comparison with previous state-of-the-art methods on RHP. Following~\cite{moon2020interhand2}, the checkmark denotes a method use ground-truth information during inference time. S and H denote scale and handness, respectively.}
\vspace{-12pt}
\label{tab:rhp}
\end{table}

\textbf{RHP dataset}:
Comparison on RHP dataset is shown in Table~\ref{tab:rhp}.
It shows that A2J-Transformer outperforms previous methods without relying on ground-truth information during inference time.
% Our method combines A2J with Transformer to effectively recover not only local details, but also to capture global information of the input images.
The experiments demonstrate the effectiveness on in-the-wild images and shows the good generalization ability of A2J-Transformer.

\begin{table}
\scriptsize
\centering
\begin{tabular}{lcc}%四个c代表有四列且内容居中
\toprule%第一道横线
Methods  &  Mean Error (mm)  &  FPS(s)  \\
\midrule%第二道横线 
Moon \etal~\cite{moon2018v2v}   &  9.22   &  35      \\
Xiong \etal~\cite{xiong2019a2j} &  8.61   &  105.06  \\
Fang \etal~\cite{fang2020jgr}   &  \textbf{8.29}   & \textbf{111.20} \\
Ours   &   8.43   &  24.81  \\
\bottomrule%第四道横线
\end{tabular}
\vspace{-4pt}
\caption{Performance comparison on NYU dataset~\cite{tompson2014real}. Our proposed A2J-Transformer can guarantee a competitive performance for the depth image input.}
\label{tab:nyu}
\vspace{-4pt}
\end{table}

\begin{table}
\scriptsize
\centering
\begin{tabular}{lcc}%四个c代表有四列且内容居中
\toprule%第一道横线
Methods  &  Mean Error (mm)  &  FPS(s)  \\
\midrule%第二道横线 
Ge \etal~\cite{ge2018hand}      &  11.30  &  48  \\
Yuan \etal~\cite{yuan2018depth} &  9.97   &  -   \\
Moon \etal~\cite{moon2018v2v}   &  9.95   &  3.5 \\
Xiong \etal~\cite{xiong2019a2j} &  8.57   & \textbf{105.06}\\
Ours   &  \textbf{8.32}   &  24.81  \\
\bottomrule%第四道横线
\end{tabular}
\vspace{-4pt}
\caption{Performance comparison on HANDS 2017 dataset~\cite{yuan20172017}. Our method can get state-of-the-art performance on this dataset.}
\label{tab:hands2017}
\vspace{-12pt}
\end{table}

\textbf{NYU and HANDS 2017 dataset}:
% We compare A2J-Transformer with state-of-the-art depth based single hand estimation methods~\cite{xiong2019a2j,moon2018v2v,fang2020jgr,ge2018hand,yuan2018depth} on NYU and HANDS 2017 dataset.
Comparison with state-of-the-art depth based single hand estimation methods on NYU and HANDS 2017 dataset are given in Table~\ref{tab:nyu} and Table~\ref{tab:hands2017}.
Since A2J-Transformer is not specifically designed for single hand estimation on depth image, we just changed the input channel to verify the generalization ability of our model through this experiment.
% The experimental results are given in Table~\ref{tab:nyu} and Table~\ref{tab:hands2017} on the average 3D distance error.
We can summarize that:

$\bullet$ Although A2J-Transformer is based on the RGB image of interacting hands, it still achieves state-of-the-art performance on HANDS 2017 dataset and gets comparable result on NYU dataset.
This relys on the strong ability of A2J-Transformer to grasp the articulated hand information and the fitting ability of 3D anchor points.
Compared with A2J~\cite{xiong2019a2j}, certain performance improvement can be achieved on two datasets, which proves that A2J-Transformer has a strong generalization ability.

\begin{figure*}[!t]
\centering
\subfloat[Weight visualization on right middle PIP.]{\includegraphics[height=4.1cm]{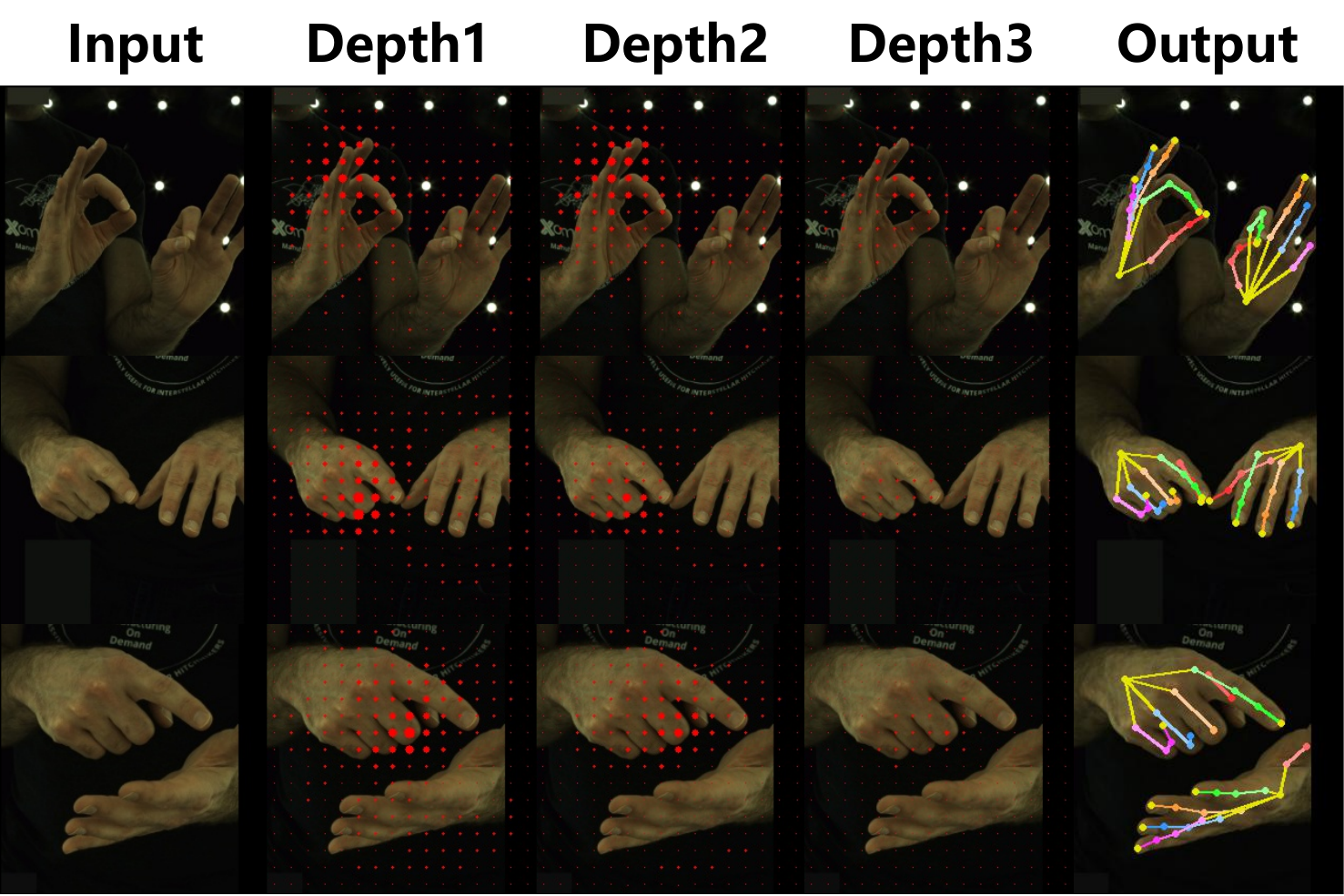}%
\label{fig:sampled_feature}}\hspace{0.4cm}
\subfloat[Weight visualization on different joints.]{\includegraphics[height=4.1cm]{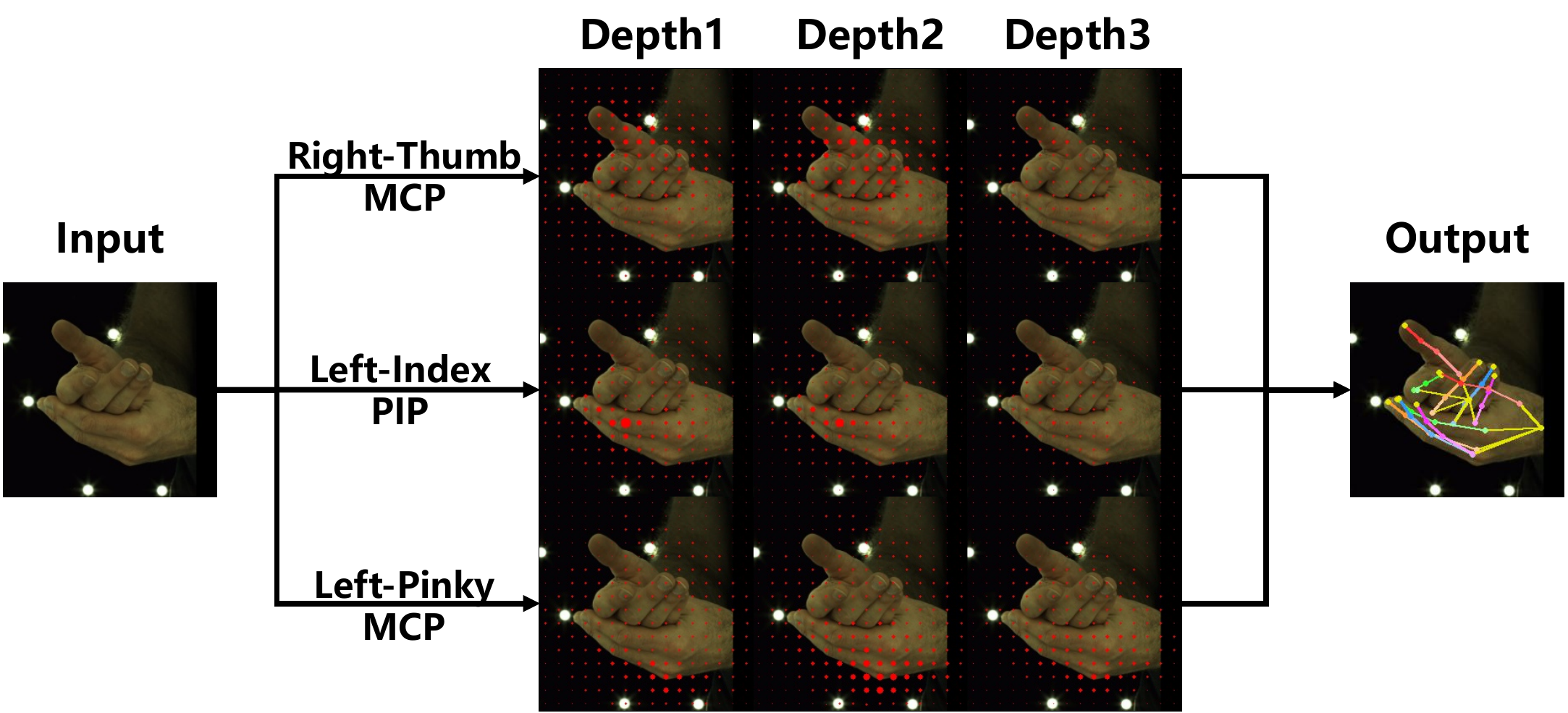}
\label{fig:ground_truth}}
\vspace{-4pt}
\caption{Qualitative results of A2J-Transformer. We show the input, output and weights of anchors on different depth value layers. Red dots in the three depth maps indicate the anchors set at depth positions +100mm, 0mm, and -100mm from the root joint respectively. The shade of red dots represent the weights
assigned to these anchors as described in Sec.~\ref{Anchor offset-weight estimation model}.}
\label{fig:show}
\vspace{-12pt}
\end{figure*}

\subsection{Ablation study}
\subsubsection{Component effectiveness analysis}
\vspace{-4pt}
\ 
\indent
The component effectiveness analysis within A2J-Transfomrer is executed on Interhand2.6M dataset.
We explore the effectiveness of four parts:  (1) Transformer-based model (anchor refinement model), (2) A2J (anchor-to-joint) module, (3) 3D anchor weights, (4) MSDAM.
The specific implementation details are respectively set as: (1) replacing the anchor refinement model with the convolution modules in A2J, (2) directly regressing the hand joints without using anchor-to-joint module, (3) setting the weights of all anchors to all the same values and normalize them, (4) replacing the MSDAM with the origin attention module.
The results are listed in Table~\ref{tab:ablation}.
It can be observed that:

\begin{table}
\scriptsize
\centering
\begin{tabular}{ccccc}%四个c代表有四列且内容居中
\toprule%第一道横线
Trans. &  A2J  &  Weights  &  MSDAM  &  MPJPE (mm)   \\
\midrule%第二道横线 
 \XSolidBrush   & \checkmark   & \checkmark   & \checkmark   &  14.44  \\
 \checkmark   & \XSolidBrush   & \checkmark   & \checkmark   &  15.36  \\
 \checkmark   & \checkmark   & \XSolidBrush   & \checkmark   &  14.04  \\
 \checkmark   & \checkmark   & \checkmark   & \XSolidBrush   &  10.69  \\
 \checkmark   & \checkmark   & \checkmark   & \checkmark     &  \textbf{9.63}  \\
\bottomrule%第四道横线
\end{tabular}
\vspace{-4pt}
\caption{Component effectiveness analysis of A2J-Transformer. `Trans.' means Transformer-based model (anchor refinement model) and `Weights' means 3D anchor weights.}
\label{tab:ablation}
\vspace{-12pt}
\end{table}

$\bullet$ After removing the Transformer-based model and A2J module, the performance of A2J-Transformer drops by 5 mm and 6 mm respectively, proving the effectiveness of addressing 3D interacting hand pose estimation task via concerning local and global visual context jointly.

$\bullet$ After removing the 3D anchor weights, the performance of A2J-Transformer drops by 4.4 mm.
This proves that there is a performance difference in the regression results of each 3D anchor point, so the weights predicted by the model are crucial for the prediction of hand joints.

$\bullet$ After replacing the MADAM with the origin attention module, the performance of A2J-Transformer drops by 1mm, which proves the MSDAM is useful to our model.

\vspace{-12pt}
\subsubsection{Anchor setting analysis}
\ 
\indent
\vspace{-14pt}

% In A2J-Transformer, we adapt 3D anchors with 3 depth values and set an anchor point every 16 pixels in the plane range.
In order to explore the impact on model performance, more in-plane and depth values are set for comparative experiments.
The specific setting methods and their performance results are shown in Table.~\ref{tab:ablation2}.
All depth values are uniformly selected near the hand joints, just like the selection of 3 depth values as described in Sec.~\ref{Anchor offset-weight estimation model}.
It can be noticed that, when more anchor in-plane and depth values are set, the performance of A2J-Transformer will improve while the inference speed will decrease in general.
In order to strike a balance between accuracy and efficiency, the value of 256 and 3 are finally choosen.

\begin{table}
\scriptsize
\centering
\begin{tabular}{cccc}%四个c代表有四列且内容居中
\toprule%第一道横线
In-plane  &  Depth  &   MPJPE (mm)  & FPS (s) \\
\midrule%第二道横线 
256  &  7  &  9.50   &  19.33 \\
256  &  5  &  9.61   &  21.21 \\
256  &  3  &  9.63   &  25.65 \\
256  &  1  &  9.75   &  26.06 \\
64   &  3  &  12.28  &  25.25 \\
16   &  3  &  14.07  &  27.39 \\
4    &  3  &  15.48  &  27.63 \\
\bottomrule%第四道横线
\end{tabular}
\vspace{-4pt}
\caption{Anchor setting analysis of A2J-Transformer. `In-plane' and `Depth' denotes the number of selected anchor number values for in-plane and depth direction, respectively.}
\label{tab:ablation2}
\vspace{-12pt}
\end{table}

\subsection{Qualitative evaluation and limitation}
We show the qualitative evaluation results in Fig.~\ref{fig:show}.
We can see that, A2J-Transformer could automatically enlarge the informative anchors' weights when different joint coordinates need to be predicted.
The model achieves accurate results even with severe occlusions in the interacting hands.
The major limitation of our method is when there is a large area of occlusion or missing in the hand area, the results predicted by our model will have deviations.

% \subsection{Limitation}
% The major limitation of our method is when there is a large area of occlusion or missing in the hand area, the results predicted by our model will have deviations.

%---------------------------------------------------------------------------------------
\section{Conclusion}
% In this paper, an anchor-transformer based 3D hand pose estimation approach for interacting hand RGB image termed A2J-Transformer is proposed.
In this paper, an 3D monocular RGB interacting hand pose estimation approach termed A2J-Transformer is proposed.
Equipped with Transformer’s non-local encoding-decoding framework, A2J is evolved to capture interacting hands’ local fine details and global articulated clues among joints simultaneously.
Besides, 3D anchors are used to better fit the depth information
and estimation of accurate 3D coordinates.
Experiments on InterHand2.6M and RHP dataset demonstrate the effectiveness and superiority of A2J-Transformer and extensions on NYU and HANDS 2017 dataset show the generalization ability.
% The major limitation of our method is when there is a large area of occlusion or missing in the hand area, the results predicted by our model will have deviations.
In future work, we will try to represent the movement of anchor points and extend our method to model-based region.

\section*{Acknowledgment}
This work is jointly supported by the National Natural Science Foundation of China (Grant No. 62271221 and U1913602). Joey Tianyi Zhou is funded by the SERC (Science and Engineering Research Council) Central Research Fund (Use-Inspired Basic Research), and the Singapore Government's Research, and Innovation and Enterprise 2020 Plan (Advanced Manufacturing and Engineering Domain) under programmatic Grant A18A1b0045.

%%%%%%%%% REFERENCES
{\small
\bibliographystyle{ieee_fullname}
\bibliography{A2J-Transformer}
}

\end{document}